\documentclass[10pt,twocolumn,letterpaper]{article}

\usepackage{iccv}
\usepackage{times}
\usepackage{epsfig}
\usepackage{graphicx}
\usepackage{amsmath}
\usepackage{amssymb}

\graphicspath{{./Figures/}}
\newcommand{\ig}[2]{\includegraphics[width=#1]{#2} } 
\newlength{\sm}
\usepackage{multirow}

\usepackage{adjustbox}
\usepackage{array}
\newcolumntype{R}[2]{%
  >{\adjustbox{valign=m, angle=#1,lap=\width-(#2)}\bgroup}%
  l%
  <{\egroup}%
}
\newcommand*\rot{\multicolumn{1}{R{90}{1em}}}% no optional argument here, please!

\usepackage[pagebackref=true,breaklinks=true,letterpaper=true,colorlinks,bookmarks=false]{hyperref}

\iccvfinalcopy % *** Uncomment this line for the final submission

\def\iccvPaperID{****} % *** Enter the ICCV Paper ID here

\ificcvfinal\fi
\begin{document}

%%%%%%%%% TITLE
\title{Can the early human visual system compete with Deep Neural Networks?}

\author{Samuel Dodge and Lina Karam\\
Arizona State University\\
{\tt\small \{sfdodge,karam\}@asu.edu}
}

\maketitle

\begin{abstract}
  We study and compare the human visual system and state-of-the-art deep neural networks on classification of distorted images. Different from previous works, we limit the display time to 100ms to test only the early mechanisms of the human visual system, without allowing time for any eye movements or other higher level processes. Our findings show that the human visual system still outperforms modern deep neural networks under blurry and noisy images. These findings motivate future research into developing more robust deep networks.
\end{abstract}

\section{Introduction}

% basic introduction
It has been shown that state-of-the-art deep neural networks (DNNs) often achieve better performance compared with human subjects on large scale classification tasks \cite{imagenet}. Given this, we might conclude that the DNNs do a better job of representing and organizing visual data. But does the success of DNNs carry over to other more difficult tasks?

% what the hell is this about
%% Deep neural networks are a sort of ``black box''. The internal mechanisms of learned DNNs are not easily explained. Furthermore, DNNs exhibit the unusual phenomenom of adversarial samples, where similar input images can be slightly perturbed to yield very different DNN outputs \cite{adversarial2}.

% distortion
Recently it has been discovered that deep neural networks perform poorly in the presence of distortions such as blur and noise \cite{dodge, bmvc}. Blur removes high frequency information and likewise, noise injects high frequency information. Current deep networks seem to experience difficulty reasoning in the presence of high levels of such distortions.

% leading to humans
Do humans have a similar trouble with distorted images? By studying the human visual system (HVS), we can perhaps gain insight into how to build DNN models that are more robust to distortions. If human performance on distorted images is better than DNNs, then this exposes a vulnerability in the DNN's representation of visual data. But if human performance is also poor, then recognition under distortions may be inherently difficult.

% what is unique about this study
Previous studies have tested human capability for recognition under noise and blur \cite{tinyimages, facenoise}, finding that humans have some robustness with respect to these distortions. In this work, we wish to compare human ability and DNN ability on a common task. Recent works perform similar experiments with distorted images \cite{sam, concurrent}. However, in this work we wish to test the \emph{early human vision system} by limiting the stimuli display time to 100ms. Within 100ms, there is no time for eye movements \cite{eye-movements}, thus the human visual system is limited to more global ``gist'' representations \cite{gist}. Can the human visual system still recognize distorted images only by the gist? By contrast, the experiments in \cite{sam} allow the subject to view the image for unlimited duration. This allows the subject to analyze more local information that can help classify the stimuli.
%The experiments in this work attempt to answer whether the gist or local analysis by eye movements is more important for recognition of distorted visual stimuli.

%% This motivates DNNs to either focus on incorporating attention mechanisms (e.g.,~\cite{attention}) or explore other mechanisms from the human visual system (e.g., feedback~\cite{feedback}).

% kind of final conclusion about usefullness of study?
%% For a vision system (human or artificial) to succesfully classify distorted images it must rely more on global inference, instead of local textures. The level of distortion can control the difficulty of the problem. Thus we can create artificial datasets with controllable difficulty. From this we can see under what conditions does the system ``break down''. The human system breaks down at a higher level of distortion than modern artificial vision systems, and thus we can take motiviation from the human visual system to create more robust artificial vision systems.

% some kind of bridge?

% Perhaps deep neural networks rely too much on accurate local filters. If the response of these local filters in early layers is incorrect then there is a cascading error which is difficult to correct. Networks with feedback (e.g. \cite{}) offer one potential 

\subsection{Related Works}
% start with human performance
Human performance on distorted stimuli has been extensively studied. Torralba \etal \cite{tinyimages} showed that humans are able to recognize very low resolution images. Similarly, studies on face images show that the human visual system can perform well in the presence of blur \cite{faceblur} and noise \cite{facenoise}.

% move to computer
There are also several works that study deep neural network performance on distorted data. Dodge and Karam~\cite{dodge} studied several different types of common image distortions and found that noise and blur have the largest effect on the performance of DNNs on the ImageNet dataset. Rodner \etal show similar findings on smaller fine grained datasets~\cite{bmvc}.

% talk about monkey studies
% also should steal some relevant things from this paper
%% The human visual system and DNNs show similarity in responses. Cadieu \etal \cite{primate-100ms} compare deep neural network response and the responses of macaque visual system. Similar to our experiment, a 100ms display time is used. The experiments show that deep networks can explain some of the response of biological vision systems. However there is still a representation gap between human neural responses and DNNs. 

% borji
Comparing human and machine vision performance has also been studied in the past. Borji and Itti \cite{borji-human} compare 14 different computer vision models on several datasets and compare with human performance. However the study does not consider modern neural networks, which greatly outperform older vision models.

Fleuret \etal compare human and machine vision performance on synthetic visual reasoning tasks \cite{fleuret}. These tasks are designed such that there must be some sort of reasoning, instead of pure pattern recognition. For many of these tasks humans outperform artificial vision systems. A followup study showed that, for some of these synthetic problems, state-of-the-art neural networks surprisingly achieve accuracy equivalent with random chance \cite{25years}.

% rewrite
Parikh \cite{parikh} studies human and artificial vision system performance on jumbled images. These jumbled images are formed by randomly permuting blocks of an image. On these jumbled images human classification performance is degraded to near the performance of a bag-of-words based classifier. While the jumbled images may give some insight into the human visual system, jumbled images are not typical visual stimuli.

%% The results show that humans and vision algorithms perform similarly for jumbled images. This shows that, at a local level, both vision systems can perform well. But at a global level there may still be a gap between human and machine performance. Our study is related, in that global information is needed to recognize highly distorted images.

% that paper
Kheradpisheh \etal \cite{deep-human} compare human and DNN performance on images of objects with arbitrary backgrounds and rotations. The highest performing DNNs match human performance. This is consistent with other studies (e.g.,~\cite{imagenet}) that show that DNN classification performance is at-par with or superior to human performance.

% cite my previous paper
To further evaluate human vs. DNNs for classification we design our experiments to test highly distorted images. We extend on the work in \cite{sam}, where Amazon Mechanical Turk testing is used to compare human and machine performance on a subset of the ImageNet dataset with added distortions. In this work there are two primary differences. First we use human subjects in a controlled lab setting. This is in contrast to Amazon Mechanical Turk studies where there is no mechanism to control viewing distance, screen brightness, etc. Secondly, we use a fixed display time instead of free viewing, which lets us analyze the accuracy of the HVS without allowing for higher level processes like eye movements.

% get other methods from concurrent paper
A concurrent independent study \cite{concurrent} compares human and deep learning performance on distorted images. However, the study does not fine tune networks on distorted images. This gives an unfair advantage to the human subjects, which may have previously seen distorted images. Additionally, we chose a 100ms display time which is less than the 200ms display time in \cite{concurrent}. This ensures that we are testing early ``gist''-based processes of the visual system.

%% The conclusions from the concurrent study in \cite{concurrent} and ours are largely similar: humans outperform neural networks on distorted stimuli. However note that we perform fine-tuning of networks with distorted images, whereas \cite{concurrent} considers networks trained on clean images. Fine-tuning is a more fair comparison, because humans have previous experience with blurry and/or noisy images. Additionally, fine-tuning on distorted images has been suggested as a method to add robustness to distortions to DNNs \cite{my-iccv-paper, blurnetworks}.

\section{Methods}

In this section we first introduce our dataset and image distortions. Then, we describe our experimental setup to test classification under distortion by both human subjects and deep neural networks.

% need the figure of the stimuli
\begin{figure*}[!tb]
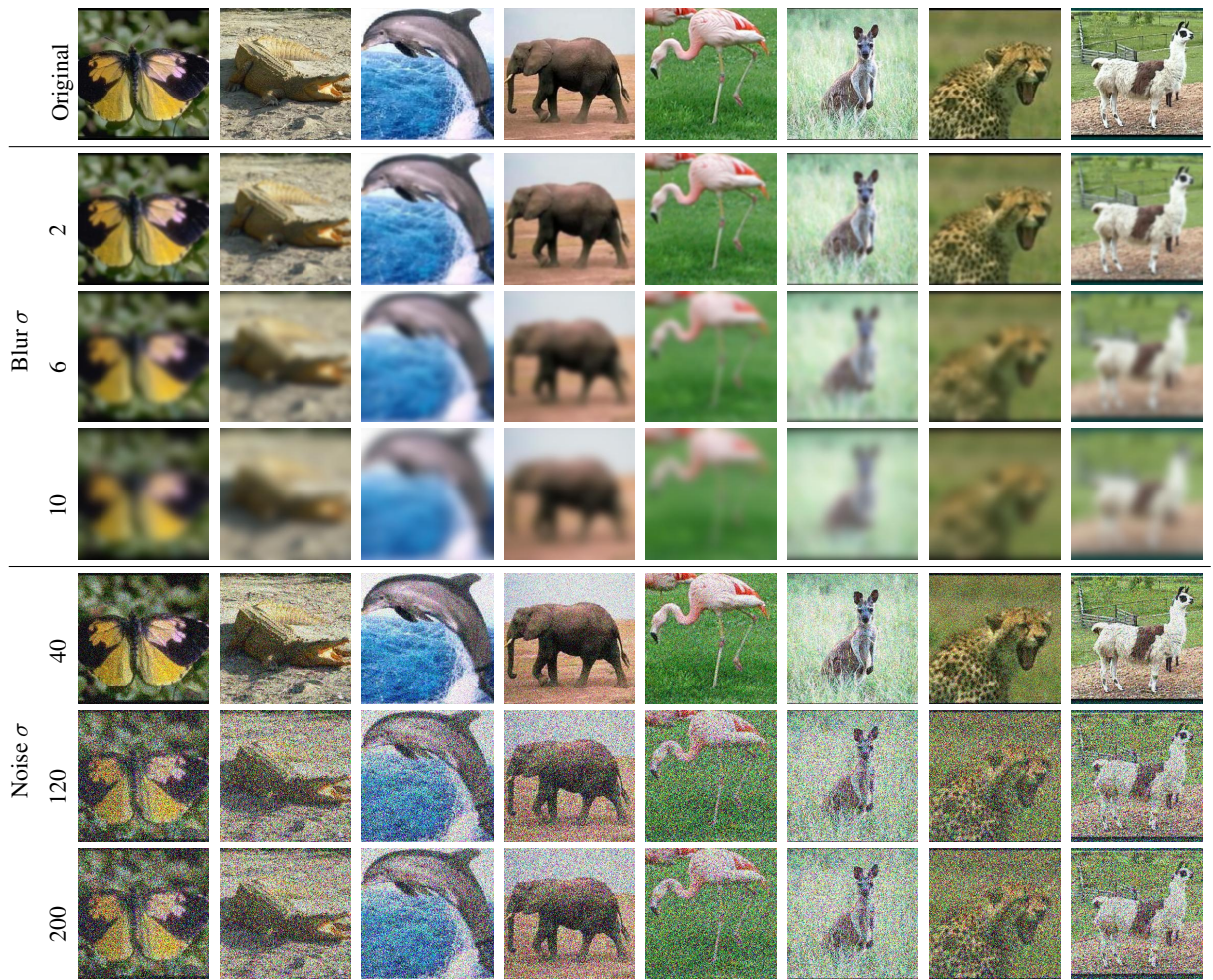

  \centering
  \small
\setlength{\tabcolsep}{1pt}
\renewcommand{\arraystretch}{0.8}
  \begin{tabular}{cccccccccccc}
   & \rot{\ \ \ Original} &
   \ig{\sm}{example_stimuli/blur0_0.jpg} & 
   \ig{\sm}{example_stimuli/blur1_0.jpg} & 
   \ig{\sm}{example_stimuli/blur2_0.jpg} & 
   \ig{\sm}{example_stimuli/blur3_0.jpg} & 
   \ig{\sm}{example_stimuli/blur4_0.jpg} & 
   \ig{\sm}{example_stimuli/blur5_0.jpg} & 
   \ig{\sm}{example_stimuli/blur6_0.jpg} & 
   \ig{\sm}{example_stimuli/blur7_0.jpg} \\
  \hline \\[-1.6ex]
  %% \multirow{4}{*}[-10pt]{\rot{Blur $\sigma$}} &
  %% \rot{\multirow{4}{*}[-10pt]{Blur $\sigma$}} &
  \parbox[t]{5mm}{\multirow{4}{*}[-10pt]{\rotatebox[origin=c]{90}{Blur $\sigma$}}} &
  \rot{\hspace{1.8em} 2} &
  \ig{\sm}{example_stimuli/blur0_2.jpg} & 
  \ig{\sm}{example_stimuli/blur1_2.jpg} & 
  \ig{\sm}{example_stimuli/blur2_2.jpg} & 
  \ig{\sm}{example_stimuli/blur3_2.jpg} & 
  \ig{\sm}{example_stimuli/blur4_2.jpg} & 
  \ig{\sm}{example_stimuli/blur5_2.jpg} & 
  \ig{\sm}{example_stimuli/blur6_2.jpg} & 
  \ig{\sm}{example_stimuli/blur7_2.jpg} \\
  &  \rot{\hspace{1.8em} 6} &
  \ig{\sm}{example_stimuli/blur0_6.jpg} & 
  \ig{\sm}{example_stimuli/blur1_6.jpg} & 
  \ig{\sm}{example_stimuli/blur2_6.jpg} & 
  \ig{\sm}{example_stimuli/blur3_6.jpg} & 
  \ig{\sm}{example_stimuli/blur4_6.jpg} & 
  \ig{\sm}{example_stimuli/blur5_6.jpg} & 
  \ig{\sm}{example_stimuli/blur6_6.jpg} & 
  \ig{\sm}{example_stimuli/blur7_6.jpg} \\
  &  \rot{\hspace{1.5em} 10} &
  \ig{\sm}{example_stimuli/blur0_10.jpg} & 
  \ig{\sm}{example_stimuli/blur1_10.jpg} & 
  \ig{\sm}{example_stimuli/blur2_10.jpg} & 
  \ig{\sm}{example_stimuli/blur3_10.jpg} & 
  \ig{\sm}{example_stimuli/blur4_10.jpg} & 
  \ig{\sm}{example_stimuli/blur5_10.jpg} & 
  \ig{\sm}{example_stimuli/blur6_10.jpg} & 
  \ig{\sm}{example_stimuli/blur7_10.jpg} \\
  \hline \\[-1.6ex]
  %% \multirow{4}{*}[-10pt]{\rot{Noise $\sigma$}} &
  \parbox[t]{5mm}{\multirow{4}{*}[-10pt]{\rotatebox[origin=c]{90}{Noise $\sigma$}}} &
  \rot{\hspace{1.5em} 40} &
  \ig{\sm}{example_stimuli/noise0_40.jpg} & 
  \ig{\sm}{example_stimuli/noise1_40.jpg} & 
  \ig{\sm}{example_stimuli/noise2_40.jpg} & 
  \ig{\sm}{example_stimuli/noise3_40.jpg} & 
  \ig{\sm}{example_stimuli/noise4_40.jpg} & 
  \ig{\sm}{example_stimuli/noise5_40.jpg} & 
  \ig{\sm}{example_stimuli/noise6_40.jpg} & 
  \ig{\sm}{example_stimuli/noise7_40.jpg} \\
  &  \rot{\hspace{1.3em} 120} &
  \ig{\sm}{example_stimuli/noise0_120.jpg} & 
  \ig{\sm}{example_stimuli/noise1_120.jpg} & 
  \ig{\sm}{example_stimuli/noise2_120.jpg} & 
  \ig{\sm}{example_stimuli/noise3_120.jpg} & 
  \ig{\sm}{example_stimuli/noise4_120.jpg} & 
  \ig{\sm}{example_stimuli/noise5_120.jpg} & 
  \ig{\sm}{example_stimuli/noise6_120.jpg} & 
  \ig{\sm}{example_stimuli/noise7_120.jpg} \\
  &  \rot{\hspace{1.3em} 200} &
  \ig{\sm}{example_stimuli/noise0_120.jpg} & 
  \ig{\sm}{example_stimuli/noise1_120.jpg} & 
  \ig{\sm}{example_stimuli/noise2_120.jpg} & 
  \ig{\sm}{example_stimuli/noise3_120.jpg} & 
  \ig{\sm}{example_stimuli/noise4_120.jpg} & 
  \ig{\sm}{example_stimuli/noise5_120.jpg} & 
  \ig{\sm}{example_stimuli/noise6_120.jpg} & 
  \ig{\sm}{example_stimuli/noise7_120.jpg} \\
    \end{tabular}
  \caption{\textbf{Example image stimuli.} Our image categories consist of ``coarse'' classes that are easily recognizable under no distortion. In this figure we show three levels of distortions among the 5 levels used in the experiments.}
  \label{fig:stimuli}
\end{figure*}

\subsection{Dataset}

% describe dataset
In this paper we wish to test the early vision system. Thus we chose a simple coarse-grained dataset, instead of a more difficult fine-grained dataset (as in \cite{sam}). We select 8 classes from the Caltech101 dataset \cite{caltech101}: Butterfly, Crocodile, Dolphin, Elephant, Flamingo, Leopard, and Llama. This dataset was deliberately chosen to be ``easy'' such that under no distortion, neither the human subjects nor the deep neural networks would have any difficulty in classification.

% describe distortion
We consider two types of distortion: random additive Gaussian noise and Gaussian blur. For Gaussian noise we consider noise with a standard deviation ranging from 0 (clean images) to 200 (highly corrupted images). Similarly, for blur we consider Gaussian blur with a kernel of standard deviation that ranges from 0 to 10. Figure \ref{fig:stimuli} shows example stimuli of the 8 classes for 3 difference levels of each distortion.

% specifics of the data distribution (train, test)
Our dataset consists of 200 clean training images (25 per class), and 40 clean validation images (5 per class). We use 80 unique images for testing blur at 5 possible levels for a total of 400 images. Similarly there are 80 unique images for testing noise at 5 levels for a total of 400 images. We additionally randomly distort the training images as explained in Section 2.3. The same training, testing, and validation split was used for each human trial, as well as for the deep neural networks.

%% For each class we take 25 images for training, 5 images for validation, 10 images for testing blur, and 10 separate images for testing noise. In total this gives 200 images for training, 40 for validation, and 160 for testing.

\subsection{Human Experiments}

%% We design our human experiments in order to test the early human visual system.
The goal of our human experiments is to test the ability of the human visual system to identify images using 100ms ``gist''-level information. Other studies have shown that humans can accurately recognize images at display times below 100ms \cite{speedsight, speedsight2}. We choose 100ms because this is the same display time in the experiments in \cite{primate-100ms} which are used to correlate human and DNN neural responses.

We design our experiment to mimic the training, validation, and testing stages used for training and evaluating DNNs.

% training
As in \cite{sam} we first allow the subjects to freely view the training images. The subject must view images in all of the categories before allowed to continue with the experiment. This allows the subjects to familiarize themselves with the image categories. This training stage is analogous to the training stage used for deep neural networks.

\begin{figure}[!tb]
  \centering
  \includegraphics[width=0.45\textwidth]{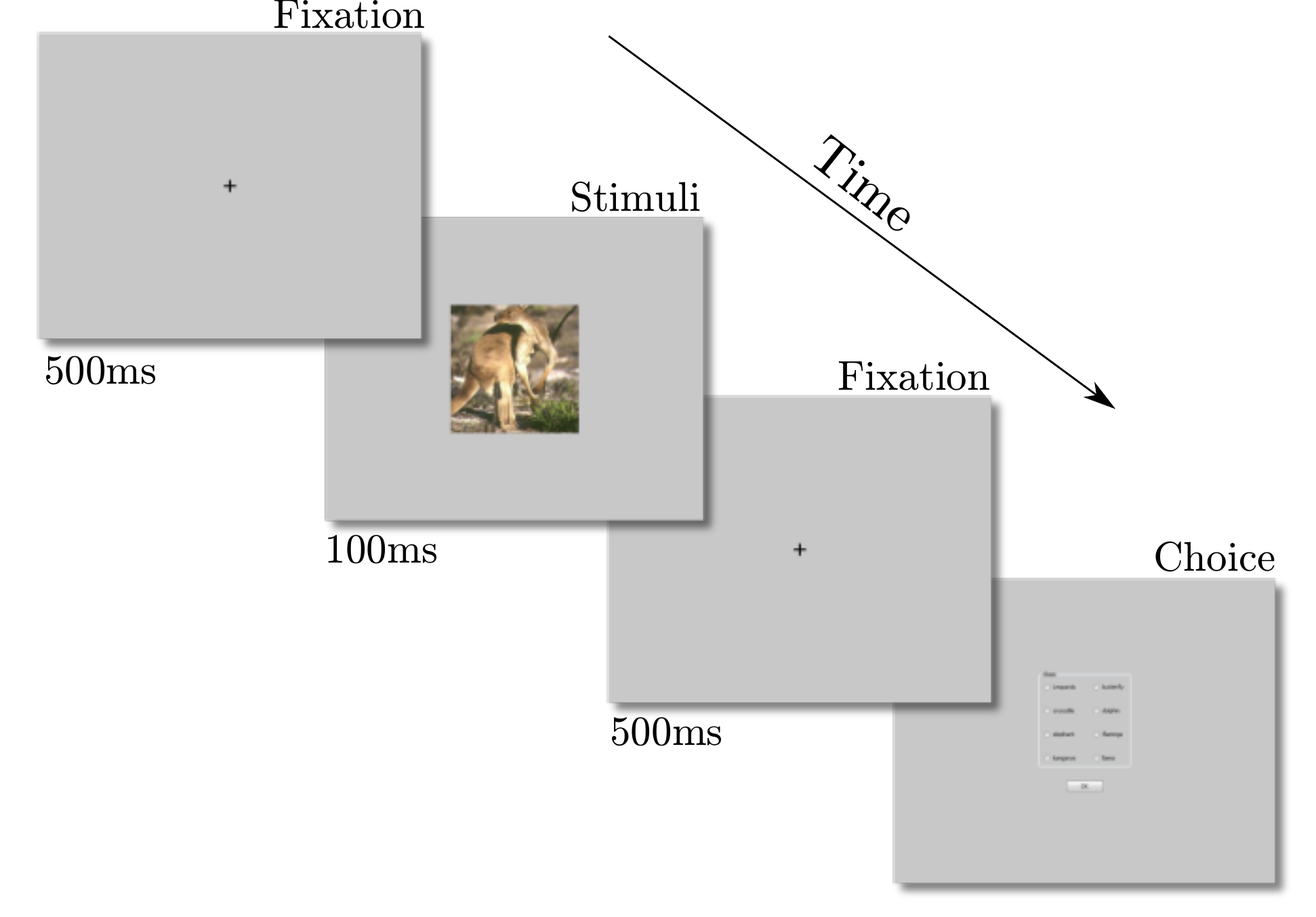}
  \caption{\textbf{Timing of validation and testing stages.} The subject is first shown a central fixation marker for 500ms followed by the image stimuli for 100ms, another fixation marker for 500ms and finally a choice screen for the subject to input the class estimate.}
  \label{fig:timing}
\end{figure}

% validation
Before we test the subject on distorted images, we first test clean images in a validation stage. This stage is to ensure that the subjects are able to correctly classify clean images. As in the rest of our experiments a central fixation cross is first shown for 500 milliseconds, followed by the image stimuli for 100 milliseconds, and finally a second fixation cross for 500 milliseconds (Figure \ref{fig:timing}). Next, a choice screen allows the subject to choose the most appropriate of the 8 categories and continue to the next image. This is a forced-choice response, so the subject must choose a class, even if the subject is not perfectly clear as to the correct class.

% need the figure of the experiment (screenshot)
\begin{figure*}[!tb]
  \centering
  \includegraphics[width=0.75\textwidth]{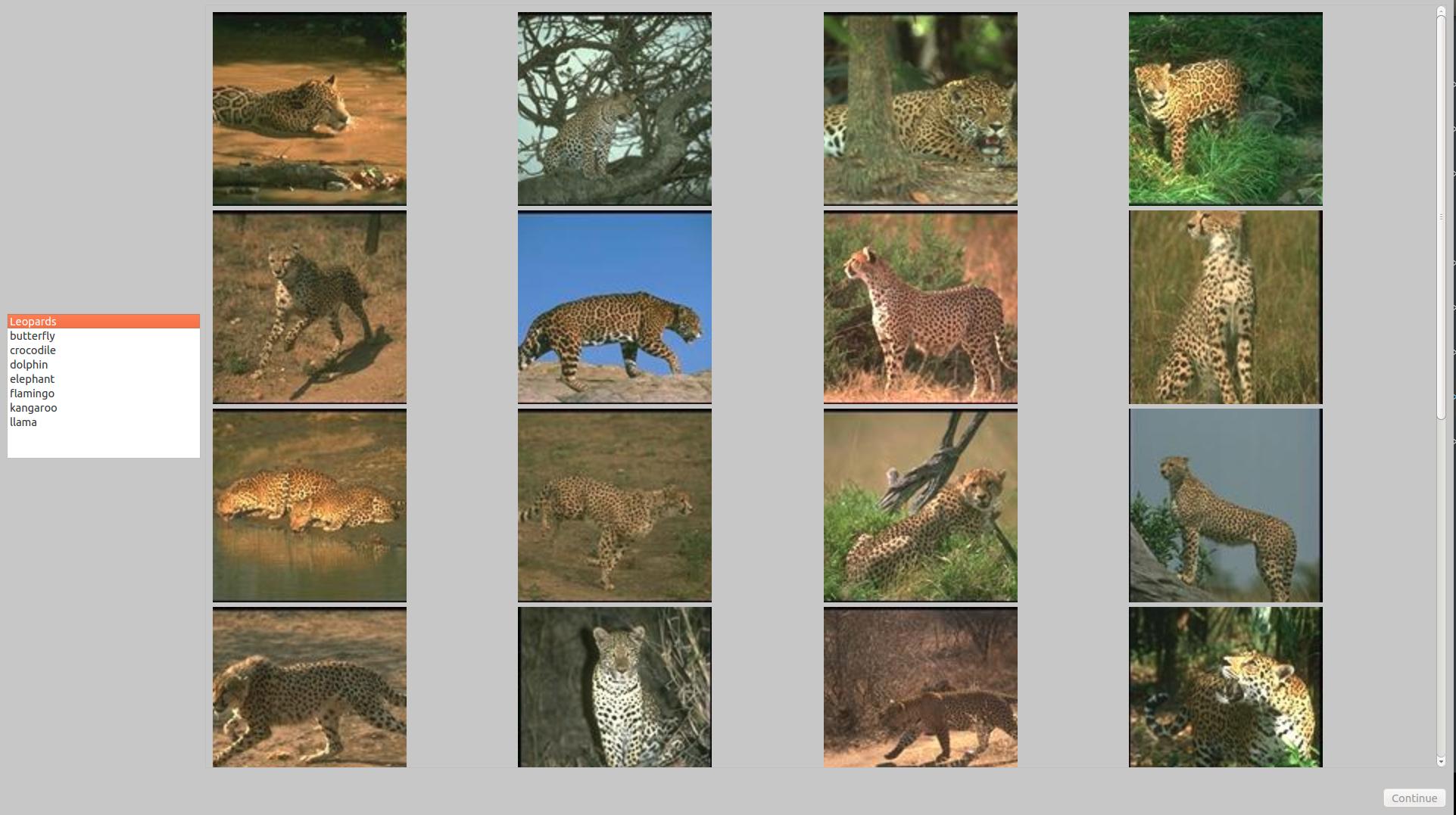}
  \caption{\textbf{Human experiment training screen.} During the training stage, the subject must view example images from each category before continuing with the experiment.}
  \label{fig:training}
\end{figure*}

% memory effect
In the next stage of the experiment the subject is asked to classify distorted images. The experiment proceeds exactly as in the validation stage, except that the images are now distorted. For each image we wish to test the maximum distortion level at which the subject can make a correct decision. Testing every possible distortion level for every image can accomplish this but has practical limitations. First, with many distortion levels, the total duration of the experiment becomes prohibitive. Secondly, multiple exposure to the same image can induce a memory effect that could help subjects identify highly distorted images that they would otherwise not be able to identify.

Instead we follow the procedure in \cite{sam}, which makes the assumption that subjects can correctly classify an image with a certain distortion level if the subject has already correctly classified an image with a higher distortion level. Thus we begin with each stimuli image at the maximum distortion level. If the subject incorrectly identifies the image, then the distortion level is reduced and the image is randomly shown later. If the subject correctly classifies an image at a particular distortion level, then all of the lower distortion levels are assumed to be correctly classified and the lower distortion levels are not tested.

% about human subjects (also can have some bullshit about monitor setup etc)
We recruit 8 subjects to participate in our experiment. All of the subjects were tested for vision and color blindness before beginning the experiment.

\subsection{Deep Neural Networks}

% introduction
Deep networks are the current state-of-the-art approaches for many image recognition tasks. These networks consist of layers of convolutional filtering, pooling, and nonlinear operations. The parameters of the layers can be learned by fitting to a training set using gradient descent based optimization algorithms.

% describe models (alexnet, vgg16, google, resnet)
We consider three classification models: VGG16 \cite{vgg}, Googlenet \cite{google}, and ResNet50 \cite{resnet}. The VGG16 network is a popular architecture which consists of 16 layers and uses convolution layers with small 3x3 filters. The Googlenet architecture introduces blocks that perform parallel convolution with filters of varying size. ResNets employ skip connections, which aids in training and can yield a more accurate network.

% distorting for fine-tuning
Each network has been pre-trained on the ImageNet dataset \cite{imagenet}. We perform fine-tuning to adapt the model to our new dataset. Specifically, we replace the last fully connected layer with a new fully connected layer with 8 units (corresponding to the 8 classes in the dataset). The learning rate of the new layer is 10x the learning rate of the pre-trained layers. The network is fine-tuned using stochastic gradient descent with momentum. We stop learning when the performance on the validation set plateaus.

% finetuning for distoritons
We test two scenarios for training. In the first scenario, the networks are fine-tuned on clean images as previously described. In the second scenario, the networks are fine-tuned on a mixture of clean images and distorted images. We refer to these networks as ``distortion-tuned''. During each mini-batch, half of the training images remain undistorted and the other half are distorted with a random level uniformly chosen from the minimum distortion level to the maximum distortion level. This procedure is identical to that in \cite{sam}, and ensures that the distortion-tuned networks can perform well at all levels of distortions, as well as for clean images.

% talk about testing
We test using the same procedure as in the human experiments. The network is first tested on the highest distortion level, and if the prediction is correct the network is assumed to correctly classify the same image at all lower levels of distortions. If the prediction is incorrect, then the network is tested on a lower level of distortion. This process continues until the network predicts the correct class, or the level of distortion becomes 0 (i.e., a clean image).

\section{Results}

% Accuracy curves
\begin{figure*}[!tb]
  \centering
  \includegraphics[width=0.95\textwidth]{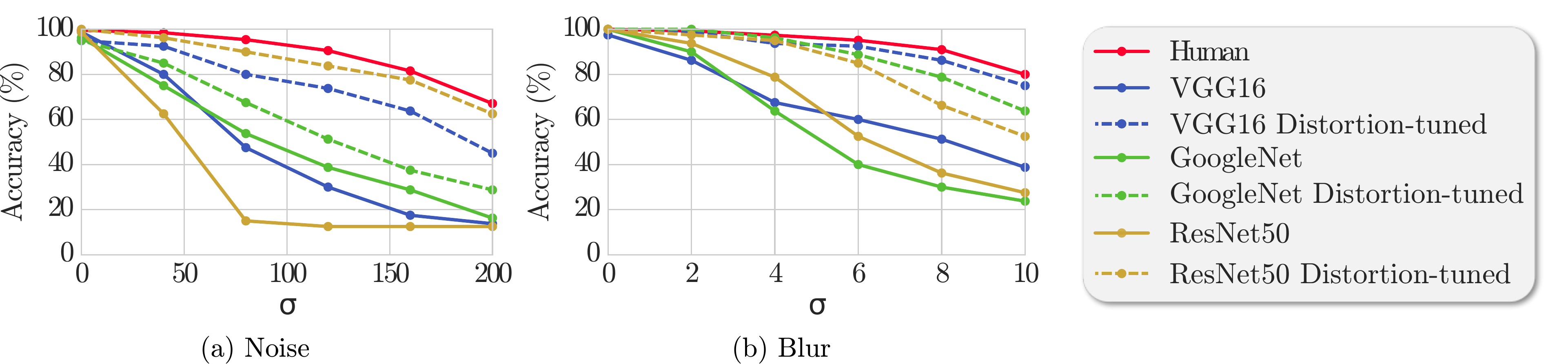}
  \caption{\textbf{Accuracy comparison between DNNs and human subjects for distorted stimuli.} Human subjects greatly outperform DNNs trained on clean data. When networks are trained on data from the respective distortion, the performance gap decreases.}
  \label{fig:accuracy}
\end{figure*}

% difference confusion matrices
\begin{figure*}[!htb]
  \centering
  \includegraphics[width=1.0\textwidth]{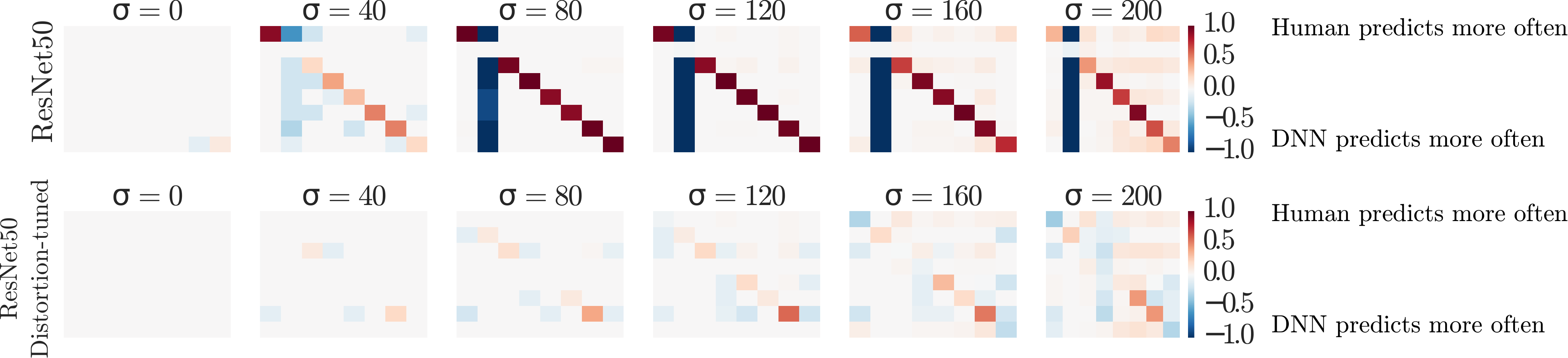}
  \caption{\textbf{Confusion matrix difference between humans and ResNet50 for noise}. Columns of a confusion matrix correspond to the actual class, and rows correspond to the predicted class. Rows are normalized to sum to one. This figure shows the \emph{difference} between the human confusion matrix and that of the ResNet50 model. Red regions denote pairs that humans predict more often, and blue regions denote pairs that networks predict more often.}
  \label{fig:noise_confus}
\end{figure*}

\begin{figure*}[!htb]
  \centering
  \includegraphics[width=1.0\textwidth]{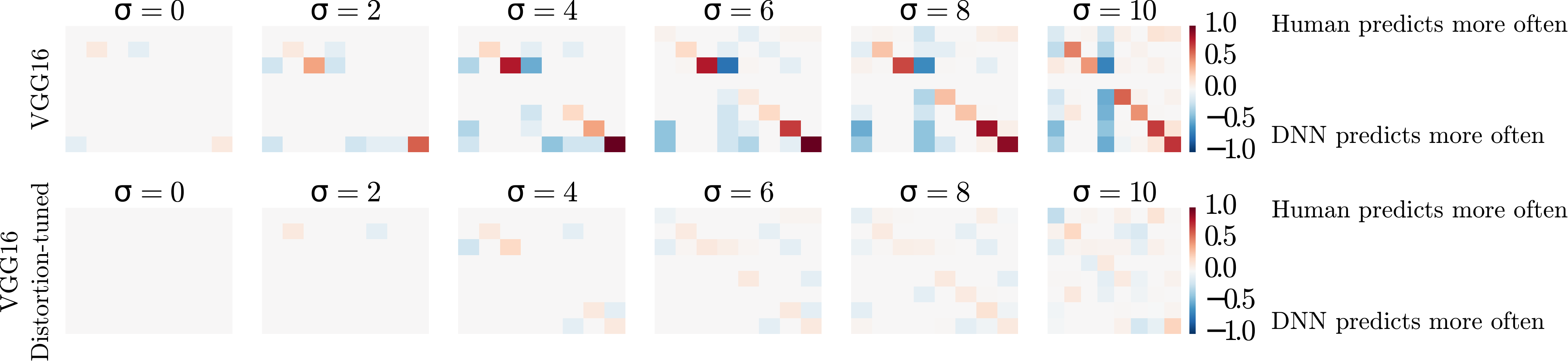}
  \caption{\textbf{Confusion matrix difference between humans and ResNet50 for blur}. Columns of a confusion matrix correspond to the actual class, and rows correspond to the predicted class. Rows are normalized to sum to one. This figure shows the \emph{difference} between the human confusion matrix and that of the VGG16 model. Red regions denote pairs that humans predict more often, and blue regions denote pairs that networks predict more often.}
  \label{fig:blur_confus}
\end{figure*}

On the validation portion of the human test, subjects scored an average accuracy of 99.3\%. This shows that the subjects are diligently performing the experiment, and that the subjects are able to classify the images correctly, even at the display time of 100ms.

% accuracy
Figure \ref{fig:accuracy} shows the comparison between human accuracy and machine accuracy for our experiments. For clean images human accuracy and DNN accuracy are nearly identical. When distortion is added, human accuracy greatly exceeds that of neural networks trained on clean images. When networks are fine-tuned on distorted data, the network performance approaches human performance for most levels of distortion. However at the largest levels of distortion, the human performance still exceeds that of the distortion-tuned deep networks.

% talk about different performance?
The network architecture can affect how resilient the network is to distortion, even when the network is distortion-tuned. For our dataset, the disortion-tuned ResNet50 model performs the best on noisy images, but is not the best performing on blurry images. For blurry images the distortion-tuned VGG16 model acheives closest to human performance.

An analysis of confusion patterns can yield insight into the behavior of the DNNs. We compute the difference between the confusion matrices of human subjects and the best performing DNN model. This tells us which categories are confused by the DNN relative to the human and vice-versa. Figure \ref{fig:noise_confus} shows this for noise using ResNet50, and Figure \ref{fig:blur_confus} shows this for blur using VGG16. For a DNN that has not seen distorted data during training, at high distortions the network tends to predict a single class regardless of the input. The network fine-tuned on distortions shows a confusion matrix that is more similar to humans.

\begin{figure}[!tb]
  \centering
  \includegraphics[width=0.45\textwidth]{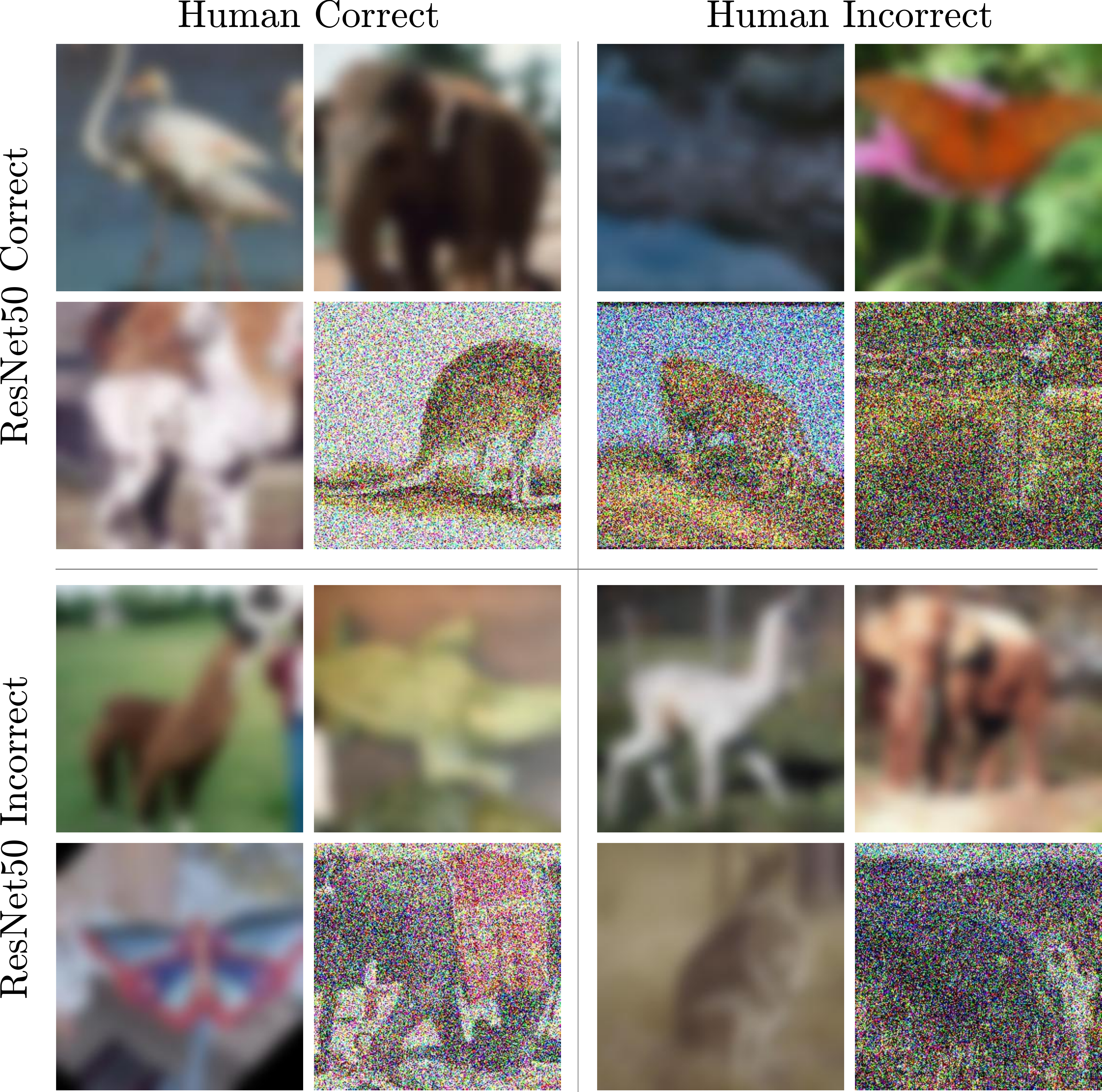}
  \caption{\textbf{Examples of correct and incorrect classifications of human subjects and distortion-tuned ResNet50 models.}}
  \label{fig:easydifficult}
\end{figure}

% difficult and easy images
We also show examples of images that are ``difficult'' or ``easy'' for humans or the distortion-tuned ResNet50 model (Figure~\ref{fig:easydifficult}). We consider an image to be correctly classified by the human subjects if 90\% or greater of the subjects selected the correct class. Some images are easy for both the neural network and human subjects. Others are consistently missclassified by both. Finally there are some images correctly classified by humans, but missclassified by DNNs. Likewise some images classified correctly by DNNs are classified incorrectly by the human subjects.

% uncertainty
%% When classifying images under high levels of distortion there is a level of uncertainty by both the human subjects and the deep neural networks. For humans it may be difficult to directly measure uncertainty, but we can use the reaction time as an indication of uncertainty. If the subject takes longer to make a decision, the subject may be more uncertain as to the correct choice. Similarly in deep neural networks a flatter probability distribution over classes is an indication of uncertainty. We use $\frac{1}{H(y)}$ as a measure of the ``flatness'' of the distribution where $H(y)$ is the entropy of the outputs ($y$) of the DNN. Figure \ref{fig:react} shows the uncertainty of the human subjects and the deep neural networks. Surprisingly, the trend of uncertainty is remarkably similar. The linear correlation between humans and DNN uncertainty is greater than 0.9 for all models and distortion types. This shows that both DNNs and human subjects experience similar changes in the level of classification uncertainty at different distortion levels.

%% % reaction time
%% \begin{figure*}[!htb]
%%   \centering
%%   \includegraphics[width=1.0\textwidth]{uncertainty.pdf}
%%   \caption{\textbf{Uncertainty in decision represented by reaction time in humans and $1/entropy$ for DNNs}. (a) shows the uncertainty as a function of noise. (b) shows the uncertainty as a function of blur. (c) shows the Pearson correlation between the uncertainty of DNNs and human subjects.}
%%   \label{fig:react}
%% \end{figure*}

\section{Discussion and Conclusion}

% provide summary of experiment and results
We performed a set of human experiments for classifying distorted images under a 100ms time constraint. We find that human performance still exceeds the performance of state-of-the-art neural networks on distorted images, even if the display time is very short (100ms). Fine-tuning with distorted images reduces this gap, but nevertheless a gap still exists. The 100ms display time does not allow for eye movements or other higher level visual processes. This tells us that the human visual system is more efficient at processing ``gist'' level information than DNNs.

% should be unique conclusion from other paper
%% Our conclusion is that the low level components of the human visual system can adequately recognize distorted images more accurately than state of the art deep neural networks.   Thus it may be useful to study the early human visual system in order to design more robust deep neural networks.

% caveat about dataset
Note that our dataset is noticeably easier than the dataset in \cite{sam}. This was to ensure that the human subjects could still perform recognition at 100ms display times. A side-effect of this choice of dataset is that it is much easier for the deep neural networks to perform recognition. For example, for such simple classes it may be sufficient to use simple color information to recognize the images. A more difficult dataset may expose a larger difference between human and DNN performance.

% takeaways for experiment
These experiments provide a look into the bias and shortcomings of DNNs by comparing them with the human visual system. Future work could analyze other scenarios where humans still outperform DNNs. These studies can help motivate future work into more robust deep neural networks.

%% One takeaway from these experiments is that we may not need to incorporate attention mechanisms (such as \cite{attention}) for image classification, since humans have high classification capability with sub-100ms ``gist''-level recognition. One significant difference between the architecture of the HVS and modern DNNs is the absence of feedback mechanisms in DNNs (although see recent work \cite{feedback}). Feedback mechanisms could enable the use of higher level information to make up for local distortions.

% nvidia acknowledgemetn
\section*{Acknowledgment}
The authors would like to thank NVIDIA Corporation for the donation of a TITAN X GPU used in these experiments.

{\small
\bibliographystyle{ieee}
\bibliography{refs}
}

\end{document}